# A State of the Art of Word Sense Induction: A Way Towards Word Sense Disambiguation for Under-Resourced Languages

Mohammad Nasiruddin
Laboratoire d'Informatique de Grenoble-Groupe d'Étude pour la Traduction Automatique/Traitement Automatisé des Langues et de la Parole
Univ. Grenoble Alpes
`mohammad.nasiruddin@imag.fr`

ABSTRACT ______________________________________________________________

Word Sense Disambiguation (WSD), the process of automatically identifying the meaning of a polysemous word in a sentence, is a fundamental task in Natural Language Processing (NLP). Progress in this approach to WSD opens up many promising developments in the field of NLP and its applications. Indeed, improvement over current performance levels could allow us to take a first step towards natural language understanding. Due to the lack of lexical resources it is sometimes difficult to perform WSD for under-resourced languages. This paper is an investigation on how to initiate research in WSD for under-resourced languages by applying Word Sense Induction (WSI) and suggests some interesting topics to focus on.

RÉSUMÉ ______________________________________________________________

### État de l'art de l'induction de sens: une voie vers la désambiguïsation lexicale pour les langues peu dotées

La désambiguïsation lexicale, le processus qui consiste à automatiquement identifier le ou les sens possible d'un mot polysémique dans un contexte donné, est une tâche fondamentale pour le Traitement Automatique des Langues (TAL). Le développement et l'amélioration des techniques de désambiguïsation lexicale ouvrent de nombreuses perspectives prometteuses pour le TAL. En effet, cela pourrait conduire à un changement paradigmatique en permettant de réaliser un premier pas vers la compréhension des langues naturelles. En raison du manque de ressources langagières, il est parfois difficile d'appliquer des techniques de désambiguïsation à des langues peu dotées. C'est pourquoi, nous nous intéressons ici, à enquêter sur comment avoir un début de recherche sur la désambiguïsation lexicale pour les langues peu dotées, en particulier en exploitant des techniques d'induction des sens de mots, ainsi que quelques suggestions de pistes intéressantes à explorer.

MOTS-CLÉS : désambiguïsation lexicale, induction de sens, langues peu dotées, ressources langagières.

## 1 Introduction

Word Sense Disambiguation (WSD) is a core and open research problem in Computational Linguistics and Natural Language Processing (NLP), which was recognized at the beginning of the scientific interest in Machine Translation (MT) and Artificial Intelligence (AI). On a variety of word types and ambiguities research has progressed steadily to the point where WSD systems achieve relatively high levels of accuracy.

The goal of a WSD is to computationally assign the correct sense of a word (i.e. meaning) in context (phrase, sentence, paragraph, text) from a predefined sense inventory, when the word has multiple meanings. It is a pervasive characteristic of natural language. The problem is that words often have more than one meaning, sometimes fairly similar and sometimes completely different. For example, the word *bank* has several senses and may refer to the edge of a river, a building, or a financial institution. The specific sense intended is determined by the textual context in which an instance of the ambiguous word appears. In "*The boy leapt from the bank into the cold water.*" the edge of a river is intended, whereas in "*The van pulled up outside the bank and three masked men got out.*" the building sense is meant, while in "*The bank sent me a letter.*" implies the financial institution sense.

Human readers have the capability to understand the meaning of a word from its context, but machines need to process textual information and transform it into data structures, which must then be analyzed in order to determine the underlying meaning. To perform WSD, a sense inventory must be available, which lists possible senses for the word of a text. A sense inventory is a lexical resource, which contains list of senses of a given word like the traditional dictionaries – knowledge resources. Manually annotated corpora with either word senses or information from knowledge sources is also an important resource for WSD.

Initially, WSD was mainly applied and developed on English texts, because of the broad availability and the prevalence of lexical resources compared to other languages. Due to the lack of lexical resources i.e. sense inventories (dictionaries, lexical databases, wordnets, etc.) and sense-tagged corpora it is difficult to start working on WSD for under-resourced languages (Bangla, Assamese, Oriya, Kannada, etc.). To account for under-resourced languages, one can easily adopt techniques aimed at the automatic discovery of word senses from text, a task called Word Sense Induction (WSI).

Languages with large amounts of data, or funding, or political interests can be interpreted as '*well-resourced*' languages, whereas, a lot of languages in the world do not enjoy this status, which is referred to in this article as '*under-resourced*' languages. This paper presents the state of the art of WSD and WSI in an under-resourced language perspective. In the following sections the paper is organized as follows: firstly, Sections 2 and 3 illustrate the main topics of interest in WSD and WSI respectively; then, Section 4 briefly describes WSI from an under-resourced language view; and finally, Section 5 concludes the article and discusses some perspectives for future work.

## 2 Word Sense Disambiguation

Depending on the degree of polysemy, there can be many different senses for a word and WSD algorithms aim at choosing the most appropriate sense combination among all possible senses for all words in a text unit (sentence, paragraph, etc.). It is essentially a task of classification for word of a text: word senses are the possible classes, the context provides evidence (features), and each occurrence of a word is assigned to one or more of its possible classes based on the evidence. There are many methods to perform WSD. In this section, various types of approaches and algorithms for WSD will be briefly presented.

The reader can refer to (Ide and Véronis, 1998) for works before 1998 and (Agirre and Edmonds, 2006), (Navigli, 2009) or (McCarthy, 2009) for a complete and current state of the art of WSD.

## 2.1 Approaches

WSD approaches can be categorized into supervised WSD and unsupervised WSD, and a further distinction can be made between knowledge-rich and knowledge-poor approaches (Navigli, 2009). Knowledge-rich methods involve the use of external knowledge sources whereas knowledge-poor methods do not. Based on machine learning techniques, researchers distinguish between supervised methods and unsupervised methods. There are three mainstream approaches to WSD, namely: Supervised WSD, Minimally-supervised WSD, and Unsupervised WSD. Figure 1 presents various WSD methods according to two axis: the quantity of annotated corpora required vertically and the amount of static knowledge horizontally.

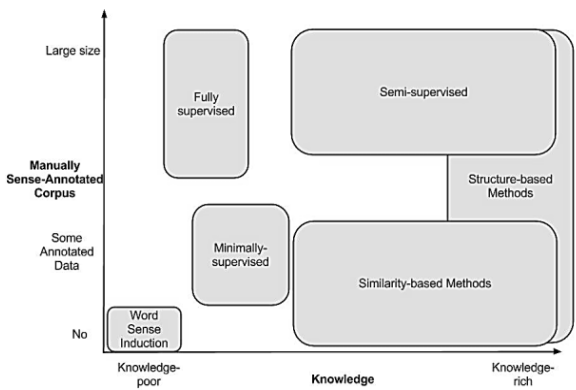

FIGURE 1 – Word Sense Disambiguation systems – Data versus Knowledge (Schwab, 2013 [personal notes]).

### 2.1.1 Supervised WSD

Supervised WSD uses supervised machine learning techniques. These approaches use a set of manually labeled training examples (i.e., sets of examples encoded as vectors whose elements represent features) to train a classifier for each target word. Support Vector Machines (SVMs), and memory-based learning have been shown to be the most successful approaches (Hoste *et al.*, 2002; Decadt *et al.*, 2004; Mihalcea *et al.*, 2004; Grozea, 2004; Chan *et al.*, 2007; Zhong and Ng, 2010), to date, probably because they can cope with the high-dimensionality of the feature space.

Supervised algorithms are progressively losing ground to the other methods. Moreover, they cannot easily be adapted to other languages without retraining (requires annotated data from that particular language). Furthermore, reusing models from one language for another leads at best to a poor classification performance (Khapra *et al.*, 2009).

### 2.1.2 Minimally-supervised WSD

Minimally supervised methods use a sense inventory, a few sense-annotated example instances, and raw corpora. From the sense-annotated examples, the system induces the senses or categories of senses for the non-annotated data, and then it functions exactly as an unsupervised clustering approach. The most prominent Minimally supervised method (Yarowsky, 1995), however, to our knowledge, has not been evaluated on SemEval WSD tasks.

#### 2.1.2.1 Knowledge-based WSD

Knowledge-based WSD algorithms are similar to Minimally-supervised WSD approaches. The objective of Knowledge-based methods is to exploit static knowledge resources, such as dictionaries, thesauri, glossaries, ontologies, collocation etc., to infer the senses of words in context. Degree (Navigli and Lapata, 2010; Ponzetto and Navigli, 2010) or Personalized PageRank (Agirre and Soroa, 2009) are among the latest knowledge-based systems in the literature that exploits WordNet (Fellbaum, 1998) or other resources like BabelNet (Navigli and Ponzetto, 2010) to build a semantic graph and use the structural properties of the graph. In order to choose the appropriate senses of words knowledge-based systems use the structural properties of the graph in context either locally to the input sentence or globally.

### 2.1.3 Unsupervised WSD

Unsupervised learning is the greatest challenge for WSD researchers. Unsupervised WSD approaches are composed of Word Sense Induction or discrimination techniques aimed at discovering senses automatically based on unlabeled corpora and then applying them for WSD. By opposition to Supervised WSD, these approaches use machine learning techniques on non-sense-tagged corpora with no *a priori* knowledge about the task at all (see Section 4).

## 2.2 Evaluation

The evaluation of previous WSD algorithms is expressed in terms of the number of "correctly" disambiguated words as evaluated through a *Gold Standard (GS)*. Since 1998, there have been several follow-up evaluation campaigns (Senseval-1 (Kilgarriff, 1998), Senseval-2 (Edmonds and Cotton, 2001), Senseval-3 (Mihalcea and Edmonds, 2004), SemEval-2007 (Navigli et al., 2007), SemEval-2010 (Agirre et al., 2010), SemEval-2012 (Manandhar and Yuret, 2012), and recently SemEval-2013 (Navigli et al., 2013)) with various disambiguation and semantic analysis tasks, which have been very successful and beneficial to the progress of the field.

In the evaluation task, a reference corpus is given with the lemmatized and part of speech (PoS) tagged instances (i.e. words), which will have to be disambiguated. The results are matched with the *GS* through *Precision (P)*, *Recall (R)*, and *$F_1$ score*, which are the standard-measures for evaluating WSD algorithms (Navigli, 2009). The evaluation tools provided calculate:

- *P*, the number of correct answers provided over the total number of answers provided;
- *R*, the number of correct answers provided over the total number of expected answers;
- *$F_1$ measure*, the harmonic mean between the two: $(2.P.R)/(P+R)$.

When all words are annotated by a WSD algorithm, then $P=R=F_1$.

## 3 Word Sense Induction

Word sense induction (WSI) is the task of automatically identifying the senses of words in texts, without the need of handcrafted resources or manually annotated data. It is an unsupervised WSD technique use machine learning methods on raw corpora without relying on any external resources such as dictionaries or sense-tagged data. During the learning phase, algorithms induce words senses from raw text by clustering word occurrences following the *Distributional Hypothesis* (Harris, 1954; Curran, 2004). This hypothesis was popularized with the phrase "*a word is characterized by the company it keeps*" (Firth, 1957). Two words are considered semantically close if they co-occur with the same neighboring words. As a result, shifting the focus away from how to select the most suitable senses from an inventory towards how to automatically discover senses from a text. By applying WSI it is possible to mitigate the *Knowledge Acquisition Bottleneck* (Wagner, 2008) problem. The single common thread to WSI methods is the reliance on clustering algorithms used on the words in the unannotated corpus. Although the role of WSI, in a disambiguation context is to build a sense inventory that can be used subsequently for WSD, therefore WSI can be considered as part of WSD. Of course, WSI can have many more applications than building sense inventories and thus WSD.

## 3.1 Approaches

WSI algorithms extract the different senses of word following two approaches – locally and globally. Local algorithms discover senses of a word per-word basis i.e. by clustering its instances in contexts according to their semantic similarity, whereas global algorithms discovers senses in a global manner i.e. by comparing and determining them from the senses of other words in a full-blown word space model (Apidianaki and Van de Cruys, 2011). Based on the type of clustering algorithms used, will be reviewed various WSI proposed in the literature in the following subsections.

### 3.1.1 Clustering Approaches

Returning to the idea of (Harris, 1954; Curran, 2004) that word meaning can be derived from context, (Pantel and Lin, 2002) discover word senses from text. The underlying hypothesis of this approach is that words are semantically similar if they appear in similar documents, within similar context windows, or in similar syntactic contexts (Van de Cruys, 2010). *Lin's algorithm* (Lin, 1998) is a prototypical example of word clustering, which is based on *syntactic dependency statistics* between words that occur in a corpus to produce sets for each discovered sense of a target word (Van de Cruys and Apidianaki, 2011). By using a similarity function, the following clustering algorithms are applied to a test set of word feature vectors (Pantel and Lin, 2002): *K-means*, *Bisecting K-means* (Steinbatch *et al.*, 2000), *Average-link*, *Buckshot*, and *UNICON* (Lin and Pantel, 2001). *Clustering By Committee (CBC)* (Pantel and Lin, 2002) also uses syntactic contexts intended for the task of sense induction, but exploits a similarity matrix to encode the similarities between words. It relies on the notion of committees to output the different senses of the word of interest. However, These approaches are hard to apply on a large scale for many domains and languages.

### 3.1.2 Extended-clustering Approaches

Considering the observation that words tend to manifest one sense per collocation (Yarowsky, 1995), (Bordag, 2006) uses *word triplets* instead of word pairs. A well-known approach to extended-clustering is the *Context-group Discrimination* algorithm (Schütze, 1998) based on large matrix computation methods. Another approach, presented by (Pinto *et al.*, 2007), attempts to improve the usability of small, narrow-domain corpora through self-term expansion. (Brody and Lapata, 2009) shows that the task of word sense induction can also be framed in a *Bayesian* context by considering contexts of ambiguous words to be samples from a multinomial distribution. There are other extended-clustering approaches, that include the *bi-gram* clustering technique proposed by (Schütze, 1998), the clustering technique using phrasal *co-occurrences* presented by (Dorow and Widdows, 2003), the technique for word clustering using a *context window* presented by (Ferret, 2004) and the method applying the *Information Bottleneck* algorithm for sense induction proposed by (Niu *et al.*, 2007). These additional clustering techniques can be broadly categorized as either improving feature selection and enriching features or introducing more effective and efficient clustering algorithms.

### 3.1.3 Graph-based Approaches

The main hypothesis of co-occurrence graphs is assuming that the semantic of a word is represented by means of co-occurrence graph, whose vertices are co-occurrences and edges are co-occurrence relations. These approaches are related to word clustering methods, where co-occurrences between words can be obtained on the basis of *grammatical* (Widdows and Dorow, 2002) or *collocational relations* (Véronis, 2004). (Klapaftis and Manandhar, 2007) propose the idea of the *Hypergraph* model for such WSI approaches. HyperLex (Véronis, 2004) is a successful graph based algorithm, based on the identification of hubs in co-occurrence graphs that have to cope with the need to tune a large number of parameters (Agirre *et al.*, 2006b). To deal with this issue several graph-based algorithms have been proposed, which are based on simple graph patterns, namely *Curvature Clustering* (Dorow *et al.*, 2005), *Squares, Triangles and Diamonds (SquaT++)* (Navigli, 2010), and *Balanced Maximum Spanning Tree Clustering (B-MST)* (Di Marco *et al.*, 2011). The patterns aim at identifying word meanings using the local structural properties of the co-occurrence graph. A randomized algorithm which partitions the graph vertices by iteratively transferring the mainstream message (i.e. word sense) to neighboring vertices proposed by (Biemann, 2006) is *Chinese Whispers*. By applying co-occurrence graph approaches, (Agirre *et al.*, 2006a; Agirre and Soroa, 2007; Korkontzelos and Manandhar, 2010) have been able to achieve state of the art performance in standard evaluation tasks. (Jurgens, 2011) reinterpret the challenge of identifying sense specific information in a co-occurrence graph as one of community detection, where a community is defined as a group of connected nodes that are more interconnected than to the rest of the graph (Fortunato, 2010). Recently, (Hope and Keller, 2013) introduced a linear time graph-based soft clustering algorithm for WSI named *MaxMax*, which obtains comparable results with those of systems based on existing, state of the art methods.

|  | **Basic Word Co-occurrence** | **Additional Features** |
|---|---|---|
| **Classical Clustering Algorithms** | Classical Clustering | Triplet Clustering<br>Self-term Expansion<br>Context Clustering<br>Translation Features |
| **Novel Algorithms for WSI** | Clustering by Committee (CBC)<br>Information Bottleneck | Hypergraph<br>Collocation Graph<br>Bayesian Model |

TABLE 1 – Overview of Techniques for Word Sense Induction (Denkowski, 2009).

*Latent Semantic Analysis (LSA)* (Landauer and Dumais, 1997; Landauer *et al.*, 1998) is currently a very popular approach to WSI that operates on word spaces (Van de Cruys and Apidianaki, 2011). *LAS* aims at finding and extract latent dimensions of meaning using *NMF (Non-negative Matrix Factorization)*, *PCA (Principal Component Analysis)* or *SVD (Singular Value Decomposition)*. The extracted latent dimensions are then used to distinguish between different senses of a target word that are in turn used to disambiguate each given instance of that word.

### 3.1.4 Translation-oriented Approaches

WSI approaches described above cover only monolingual data; in the context of Machine Translation (MT), recent work has been done to incorporate bilingual data into the sense induction task. Translation-oriented WSI approaches involve augmenting the source language context with target language equivalents. (Apidianaki, 2008) describes this process by using a bilingual corpus that has been word aligned by type and token to construct two bilingual dictionaries, where each word type is associated with its translation equivalent. The lexicon is filtered such a way that words and their translation equivalents have matching PoS tags and words appear in the translation lexicons for both directions.

## 3.2 Evaluation

The evaluation of WSI approaches is one of the key challenges for researchers. As the sense clusters derived by these algorithms may not match the actual senses defined in lexical resources like dictionaries, lexical databases, wordnets, etc., the evaluation of these algorithms needs to be carried out manually, by asking language experts to corroborate the results. However, it is hard to evaluate the results of WSI, as the resulting clustered senses can vary from algorithm to algorithm or even for various parameter values for a single algorithm, as even determining the number of clusters a difficult matter. From the very beginning of WSI, depending on different approaches researchers have developed various evaluation methodologies that can be separated into three main categories.

### 3.2.1 Supervised Evaluation

In this evaluation method, the target word corpus is divided into two parts – a testing and a training part. Firstly, the training part is used to map the automatically induced clusters to *Gold Standard (GS)* senses. In the next step, the test corpus is used to evaluate WSI approaches in a WSD (Agirre and Soroa, 2007) setting. Finally, the usual *Precision (P)* and *Recall (R)* are used to determine the quality of the resulting WSD.

### 3.2.2 Unsupervised Evaluation

In this evaluation setting, the induced senses are evaluated as clusters of examples, and compared to sets of examples that have been tagged with *sense labels from a Gold Standard*. The *V-measure* (Rosenberg and Hirschberg, 2007) is used to determine the quality of clusters by combining metrics such as the *Paired F-Score* (Manandhar et al., 2010), the *RandIndex* (Rand, 1971; Navigli, 2010) and others. They measure both the coverage and the homogeneity of a clustering output as opposed to the traditional clustering measure of *F-Score* (Zhao et al., 2005) that is most commonly used to assess the performance of WSI systems.

### 3.2.3 Evaluation as an Application

Recently, (Navigli and Crisafulli, 2010; Di Marco and Navigli, 2013) proposed to evaluate WSI approaches as part of a specific application, where WSI techniques have been shown to consistently surpass symbolic state of the art systems (See section 4.3). The evaluation of WSI and WSD systems was performed in the context of *Web search result clustering*.

## 3.3 SemEval WSI Evaluation Tasks

This section briefly describes the different SemEval workshops (from 2007 to 2013) with WSI evaluation tasks that focus on the evaluation of semantic analysis systems.

### 3.3.1 SemEval-2007 Task 2

The goal of SemEval-2007 Task 2 (Agirre and Soroa, 2007) is to allow for the comparison across sense-induction and discrimination systems, and also to compare these systems to other supervised and knowledge-based systems. This task evaluates WSI systems on 33 nouns and 65 verbs (lexical sample), where the corpus consists of texts of the *Wall Street Journal (WSJ)* corpus, and is hand-tagged with *OntoNotes* senses (Hovy *et al.*, 2006). For each tagged-word, the task consists of first identifying the senses of target words (e.g. as clusters of target word instances, co-occurring words, etc.), and secondly tagging the instances of the target word using the automatically induced clusters. This double evaluation methodology (i.e. supervised evaluation and unsupervised evaluation) has been attempted by (Agirre *et al.*, 2006a).

### 3.3.2 SemEval-2010 Task 14

SemEval-2010 Task 14 is a continuation of the WSI SemEval-2007 Task 2 with some significant changes to the evaluation setting. The main difference in this task compared to the SemEval-2007 WSI task, is that the training and testing data are treated separately, which allows for a more realistic evaluation of the clustering models. Readers may refer to (Klapaftis and Manandhar, 2013) for a detailed analysis of the SemEval-2010 WSI task evaluation result and new evaluation settings.

### 3.3.3 SemEval-2013 Task 11

For the evaluation in SemEval-2013: Task 11, WSI and WSD systems are applied to web search result clustering, where the test data consists of 100 topics (all nouns), each with a list of 64 top-ranking documents. The topics were selected from the list of ambiguous *Wikipedia* entries (i.e., those with "*disambiguation*" in the title) among queries of lengths ranging between 1 and 4 words. The 64 snippets associated with each topic were collected from the results provided by the *Google* search engine. Three annotators tagged each snippet with the most appropriate meaning from *Wikipedia*, with adjudication in the case of disagreement. For a detailed description of the SemEval-2013: Task 11 evaluations please refer to (Di Marco and Navigli, 2013; Navigli and Vannella, 2013).

## 4 WSI for Under-Resourced Languages

Given the above-mentioned difficulties, WSI is an attractive alternative to WSD, especially for under-resourced languages.

The Figure 2 shows an overview of the state of freely available resources for a certain number of representative languages and the aim of the diagram is to give the reader an idea of the current situation. Making a highly precise diagram would be prohibitively complex and furthermore, the position of the various languages must be interpreted relatively to each other rather than in an absolute manner (except for English that we placed in the top right-hand corner as a reference).

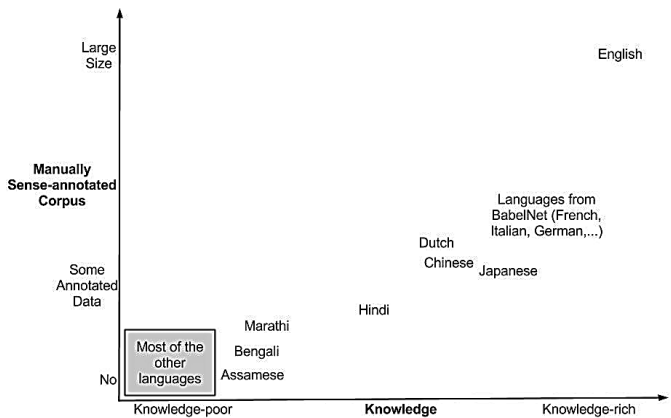

FIGURE 2 – Computationally language richness – Data versus Knowledge (Schwab, 2013 [personal notes]).

As described previously, two kinds of resources are difficult/expensive to build: annotated corpora and lexical databases (static versus dynamic knowledge). While lexical databases can be used directly in many applications or by humans, sense-annotated corpora have less applications and are much more expensive to build. An automatic procedure can extract the senses that are objectively present in a particular corpus and allows for the sense inventory to be straightforwardly adapted to a new domain. By applying WSI, it is practical to disambiguate particular word instances using the automatically extracted sense inventory. Words and contexts are mapped to a limited number of topic dimensions (depending on the topic of the words and contexts) in a latent semantic word space. A particular sense is associated with a particular topic and different senses can be discriminated through their association with particular topic dimensions. (Van de Cruys and Apidianaki, 2011) describe the induction step and disambiguation step as being based on the same principle.

Currently, *Wikipedia* contains 285 languages, anyone can generate corpus from a *Wikipedia dump*, or blogs, forums, newspaper articles in any language. As WSI is a kind of clustering problem, the evaluation of the clusters is normally difficult, however if the evaluation process is followed, it becomes rather straightforward. Though semantic evaluation campaigns are based on some dominant languages, progress for under-resourced languages is still ongoing. In this regard, Crowdsourcing (Sabou et al., 2012), especially *Games With A Purpose* (von Ahn, 2006) are considered as an attractive alternative for collecting annotated data (Wang et al., 2010), which can be subsequently used as a *GS* for evaluating the systems.

## 5 Conclusion and Discussions

The state of the art of Word Sense Disambiguation followed by Word Sense Induction techniques for under-resourced languages are provided in this article, and also tries to provide a basic idea of the essentials of the field. Here, the authors show a way to work on WSD by using WSI approaches in an unsupervised way, where very few resources (like corpora) are available. Basically, the performance of WSD systems depends heavily on which sense inventory is chosen, here WSI overcomes this issue by allowing unrestrained sets of senses. Besides, its evaluation is particularly hard because there is no easy way of comparing and ranking different representations of senses.

# Acknowledgements

The work presented in this paper was conducted in the context of the VideoSense project, funded by the French National Research Agency (ANR) under its CONTINT 2009 program (grant ANR-09-CORD-026).